\documentclass{llncs}

\usepackage[T1]{fontenc}
\usepackage[latin1]{inputenc}

\usepackage{booktabs}
\usepackage[table]{xcolor}
\usepackage{multirow}
\usepackage{soul}
\usepackage{enumitem}
\usepackage[pdftex]{graphicx}
\DeclareGraphicsExtensions{.pdf,.ai,.jpg,.png}
\setkeys{Gin}{pagebox=artbox}
\graphicspath{{./ratio24-argument-quality-figures/}}

\usepackage[hidelinks]{hyperref}

\begin{document}

\title{Are Large Language Models Reliable Argument Quality Annotators?\texorpdfstring{\thanks{Supported by the German Research Foundation (Project Nr. 455911521).}}{}}

\author{Nailia Mirzakhmedova
\and
Marcel Gohsen
\and
Chia Hao Chang
\and
Benno Stein
}

\authorrunning{Mirzakhmedova et al.}

\institute{Bauhaus-Universität Weimar, Germany \\
\email{\{first\_name, last\_name\}@uni-weimar.de}}

\maketitle

\begin{abstract}
Evaluating the quality of arguments is a crucial aspect of any system leveraging argument mining. However, it is a challenge to obtain reliable and consistent annotations regarding argument quality, as this usually requires domain-specific expertise of the annotators. Even among experts, the assessment of argument quality is often inconsistent due to the inherent subjectivity of this task. In this paper, we study the potential of using state-of-the-art large language models (LLMs) as proxies for argument quality annotators. To assess the capability of LLMs in this regard, we analyze the agreement between model, human expert, and human novice annotators based on an established taxonomy of argument quality dimensions. Our findings highlight that LLMs can produce consistent annotations, with a moderately high agreement with human experts across most of the quality dimensions. Moreover, we show that using LLMs as additional annotators can significantly improve the agreement between annotators. These results suggest that LLMs can serve as a valuable tool for automated argument quality assessment, thus streamlining and accelerating the evaluation of large argument datasets.

\end{abstract}

\section{Introduction}
\label{introduction}

Computational argumentation is an interdisciplinary research field that combines natural language processing with other disciplines such as artificial intelligence. A central question in computational argumentation is: What makes an argument good or bad? Depending on the goal of the author of a text, argument quality can involve a variety of dimensions. Evaluating the quality of an argument across these diverse dimensions demands a deep understanding of the topic at hand, often coupled with expertise from the argumentation literature. Hence, manual assessment of argument quality is a challenging and time-consuming process.

A promising technology to streamline argument quality assessment are large language models (LLMs) which have demonstrated impressive capabilities in tasks that require a profound understanding of semantic nuances and discourse structures. LLMs have been effectively employed in tasks such as summarization \cite{wang:2023b}, question answering \cite{kamalloo:2023}, and relation extraction \cite{wadhwa:2023}. Previous research has also investigated the usefulness of LLMs in argument mining tasks such as argument component identification \cite{guo:2023}, evidence detection \cite{huo:2023}, and stance classification \cite{chen:2023}. Moreover,  an emerging trend highlights the adoption of LLMs for data annotation purposes, such as sentiment analysis \cite{ding:2023,ronningstad:2024}, relevance judgement \cite{gilardi:2023}, and harm measurement \cite{magooda:2023}. To the best of our knowledge, no prior work has investigated the potential of LLMs as annotators of argument quality.

In this paper, we analyze the reliability of LLMs as argument quality annotators by comparing automatic quality judgements with human annotations from both experts and novices.\footnote{Code and data are available at \href{https://github.com/webis-de/RATIO-24}{\tt github.com/webis-de/RATIO-24}.} We compare these quality ratings not only at an aggregate level, but also examine the individual components that make up argument quality. This includes looking at how well models can judge the relevance and coherence of an argument, the sufficiency of its evidential support, and the effectiveness of its rhetorical appeal. Ultimately, our objective is to understand whether LLMs can serve as a practical and reliable tool that supports and enhances human-led effort in argument quality assessment.

Specifically, we ask the following research questions regarding the potential of employing LLMs as argument quality annotators:

\begin{enumerate}[itemindent=0em, align=left, leftmargin=3em,labelsep=0pt,labelwidth=3em]
\item[RQ1:]
Do LLMs provide more consistent evaluations of argument quality compared to human annotators?
\item[RQ2:]
Do the assessments of argument quality made by LLMs align with those made by either human experts or human novices?
\item[RQ3:]
Can integrating LLM annotations with human annotations significantly improve the resulting agreement in argument quality ratings?
\end{enumerate}

In the following, Section~\ref{related-work} reviews the related work, Section~\ref{experiment-design} describes the experimental setup, including the dataset, the annotation procedure, and the employed models, and Section~\ref{results} presents the results of these experiments.
\section{Related Work}
\label{related-work}

We first review existing literature related to the evaluation and annotation of argument quality. Following that, we explore the works that examined the capabilities of large language models (LLMs) as data annotators as well as the degree of alignment between LLMs and human annotators.

\subsection{Evaluating Argument Quality}

Collecting argument quality annotations is an intricate task that often requires domain-specific knowledge, a number of annotators, and assured consistency in annotator reliability. Numerous works have studied argumentation quality across different domains, employing multiple annotators to classify and evaluate arguments based on various quality criteria. Park and Cardie \cite{park:2014} studied argumentation quality in the domain of web discourse. They employed two annotators to classify 9,000~web-based propositions into four categories based on their level of \textit{verifiability}. Habernal and Gurevych \cite{habernal:2016} let five crowd-workers annotate a dataset consisting of 16,000 pairs of arguments with a binary \textit{``is more convincing''} label, providing explanations for their decisions. Toledo et al. \cite{toledo:2019} collected a dataset of 14,000~pairs of arguments, each annotated with relative argument quality scores ranging from 0 to~1. They employed between 15 and~17 annotators for each instance to enhance the reliability of the collected annotations.

In the domain of student essays, Persing and Vincent \cite{persing:2015} instructed six human annotators to evaluate 1,000 essays based on the \textit{strength} of argumentation on a scale from 1 to~4. Carlile et al. \cite{carlile:2018} considered \textit{persuasiveness} as the most important quality dimension of an argumentative essay. They asked two native English speakers to annotate 102~essays with argument components, argument persuasiveness scores, and further attributes such as \textit{specificity, evidence, eloquence, relevance}, and \textit{strength}, that determine the persuasiveness of an argument. Moreover, Marro et al. \cite{marro:2022} employed three expert annotators for the annotation of essay components of Stab and Gurevych \cite{stab:2017} for three basic argument quality dimensions: \textit{cogency, rhetoric}, and \textit{reasonableness}.

Aiming to create a unified understanding of argument quality properties, Wachsmuth et al. \cite{wachsmuth:2017} proposed a comprehensive taxonomy of 15~argument quality dimensions derived from the argumentation literature. Three expert annotators were employed to annotate 320~arguments \cite{habernal:2016}. In Section \ref{experiment-design}, we use their quality annotations from 1~(low) to 3~(high) as a reference for our experiments.

Despite the multiple attempts and methodologies to evaluate argument quality, the process remains labor-intensive, time-consuming, and requires a significant degree of expertise. To facilitate the task of argument quality annotation, we propose employing LLMs, as they can potentially provide more reliable and consistent annotations while significantly reducing the required manual effort.

\subsection{LLMs as Annotators}
\label{llm-as-annotators}

Recent work has expanded the role of LLMs from language generation and explored the potential of using LLMs as data annotators. Ding et al. \cite{ding:2023} assessed the performance of GPT-3 \cite{brown:2020} as a data annotator for sentiment analysis, relation extraction, named entity recognition, and aspect sentiment triplet extraction. They compared the efficiency of BERT \cite{devlin:2019}, trained using data annotated by GPT-3, against BERT trained with human-annotated data. Their findings showed a noticeably similar performance level with substantially reduced annotation costs, promising a potentially cost-effective alternative in using GPT-3 for annotation. A study by Gilardi et al. \cite{gilardi:2023} cross-examined the annotations by ChatGPT \cite{chatGPT:2022} and those by crowd-workers against expert annotations across four tasks: content relevance assessment, stance detection, topic detection, and general frame detection. They found that ChatGPT not only outperforms crowd-workers in terms of accuracy, but also shows a high degree of consistency in annotations.

The study by Gao et al. \cite{gao:2023} explored automatic human-like evaluations of text summarization using ChatGPT compared to human experts. The model was prompted to evaluate the quality of summaries based on \textit{relevance}, \textit{coherence}, \textit{fluency}, and \textit{consistency} of the generated summaries. The authors found that ChatGPT's evaluations were highly correlated with those of human experts.

Zhuo et al. \cite{zhuo:2023} proposed to use LLMs as evaluators of code generation. The authors used the CoNaLa dataset \cite{yin:2018} and reported high example-level and corpus-level Kendall-Tau, Pearson, and Spearman correlations with human-rated code usefulness for various programming languages.

In the domain of information retrieval, Faggioli et al. \cite{faggioli:2023} investigated the performance of GPT-3.5 and YouChat for query-passage relevance judgements. Given the high subjectivity of the task, their results showed a reasonable correlation between highly-trained human assessors and fully automated judgements. 

Closest to our work is that by Chiang et al. \cite{chiang:2023}, who compared the judgments of GPT-3 on text quality to expert human judgments on a 5-point Likert scale for four quality attributes: \textit{grammaticality, cohesiveness, likability}, and \textit{relevance}. Their findings revealed varying degrees of positive correlations between GPT-3 and human judgments, ranging from weak to strong. 

When compared to existing research, our work pioneers the study of argument quality annotations generated by LLMs. In order to provide a thorough evaluation, we use an inter-annotator agreement metric to assess the consistency of annotations from these models, human experts and novices. This comparison allows us to understand the alignment between LLMs and human annotators, and to determine the potential of using LLMs as argument quality annotators.
\section{Experimental Design}
\label{experiment-design}

To investigate the reliability of large language models (LLMs) as annotators of argument quality, we conduct an experiment comparing human annotations with ratings generated automatically by LLMs. We treat LLMs as separate annotators and analyze the agreement both within and across groups of humans and models.

\subsection{Expert Annotation}

\begin{table}[t]
\fontsize{8pt}{10pt}\selectfont
\setlength{\tabcolsep}{4pt}
\centering
\caption{Descriptions of argument quality dimensions as per Wachsmuth et al. \cite{wachsmuth:2017a}.}
\label{table-quality-dimensions}
\begin{tabular}{@{}l p{9cm}@{}}
\toprule
\bf Quality Dimension & \bf Description\\
\midrule
Cogency & Argument has (locally) acceptable,
relevant, and sufficient premises.\\
\ \ Local acceptability & Premises worthy of being believed.\\
\ \ Local relevance & Premises support/attack conclusion. \\
\ \ Local sufficiency & Premises enough to draw conclusion.\\
\addlinespace
Effectiveness & Argument persuades audience.\\
\ \ Credibility & Makes author worthy of credence.\\
\ \ Emotional appeal & Makes audience open to arguments.\\
\ \ Clarity & Avoids deviation from the issue, and uses
correct and unambiguous language.\\ 
\ \ Appropriatness & Language proportional to the issue,
supports credibility and emotions. \\ 
\ \ Arrangement & Argues in the right order. \\
\addlinespace
Reasonableness & Argument is (globally) acceptable,
relevant, and sufficient.\\
\ \ Global acceptability & Audience accepts use of argument.\\
\ \ Global relevance & Argument helps arrive at agreement. \\
\ \ Global sufficiency & Enough rebuttal of counterarguments.\\
\addlinespace
Overall quality & Argumentation quality in total.\\
\bottomrule
\end{tabular}
\end{table}

The goals of argumentation are manifold and include persuading audiences, resolving disputes, achieving agreement, completing inquiries, or recommending actions \cite{tindale:2007}. Due to the variety of these goals, the dimensions of argument quality are equally diverse. Based on a comprehensive survey of argumentation literature, Wachsmuth et al. \cite{wachsmuth:2017} proposed a fine-granular taxonomy of argument quality dimensions that differentiates logical, rhetorical, and dialectic aspects. An overview of all quality dimensions is provided in Table~\ref{table-quality-dimensions}. 

In their work, Wachsmuth et al. \cite{wachsmuth:2017} employed experts to rate the quality of arguments according to their proposed taxonomy. Three experts were selected out of a pool of seven based on their agreement in a pilot annotation study. These three experts comprised two PhDs and one PhD student (two female, one male) from three different countries. To construct the Dagstuhl-15512-ArgQuality corpus, the selected experts annotated 320 arguments from the UKPConvArgRank dataset \cite{habernal:2016}. The resulting corpus contains 15 quality dimensions for each argument, each rated on a 3-point Likert scale (low, medium, high) or as not assessable. Each argument in the corpus belongs to one of 16 different topics and takes a stance for or against the topic. The dataset is balanced and contains 10 supporting and 10 attacking arguments per topic. The annotation guidelines, which define all quality dimensions in more detail, are publicly available online.\footnote{\url{https://zenodo.org/records/3973285}}

\subsection{Novice Annotation}
\label{novice-annotation}

To provide an additional point of reference for determining the abilities of LLMs as argument quality annotators, we conducted an annotation study involving humans with no prior experience with computational argumentation. We asked undergraduate students to assess the quality of arguments from the Dagstuhl-15512-ArgQuality corpus using the same taxonomy.

The expert annotation guidelines require that annotators have expertise in computational argumentation. To make this task accessible for novices, we paraphrased the annotation guidelines and the definitions of argument quality dimensions to ensure clarity and comprehension. These simplified definitions for each quality dimension can be found in Appendix. To illustrate, the expert definition of \textit{local acceptability} of an argument is stated as follows:

\begin{definition}[Local Acceptability (Expert)]
A premise of an argument should be seen as acceptable if it is worthy of being believed, i.e., if you rationally think it is true or if you see no reason for not believing that it may be true.
\end{definition}

\noindent
The above definition requires an annotator to distinguish between premises and arguments. To ease the understanding and reduce the necessary prior knowledge, we simplify the definition of local acceptability as follows:

\begin{definition}[Local Acceptability (Novice)]
The reasons are individually believable: they could be true.
\end{definition} 

\noindent
We refer to arguments as ``reasons'' within the simplified guidelines and combine the stance with the issue into a ``conclusion''. For example, given the issue \textit{``Is TV better than books?''} and the stance \textit{``No it isn't''}, we state the conclusion as \textit{``TV is not better than books''}.

Each novice annotator was presented with an argument, a conclusion, and the simplified definitions of the quality dimensions. Identical to the annotation procedure for expert annotations, the annotators were tasked to rate each quality dimension of the argument on a 3-point Likert scale or as not assessable.

In total, we acquired 108 students to annotate the quality of the 320 arguments from the dataset. We assigned 10 arguments to each student to annotate in order to obtain at least three annotations per argument and quality dimension. Since not all students finished their annotations and some students annotated a wrong set of arguments, we obtained a minimum of three annotations per argument and quality dimension only for 248 arguments. We treat the missing annotations of the 72 arguments as non-evaluable. For the 163 arguments for which we collected more than 3 annotations, we select three annotations that maximize the inter-annotator agreement measured by Krippendorff's $\alpha$.

\begin{figure}[t]
\frame{
\qquad
\parbox{0.9\textwidth}{
\small\tt\raggedright
\medskip
\noindent\#\#\# Instruction: \\
Please answer the following questions for the given comment from an online debate forum on a given issue.\\[2ex]

\noindent\#\#\# Issue: \\
Is TV better than books? \\[2ex]

\noindent\#\#\# Stance: \\
No, it isn't. \\[2ex]

\noindent\#\#\# Argument: \\
Books will be always great whatever the new technological developments emerges books has its fixed place in every humans heart. \\[2ex]

\noindent\#\#\# Quality dimension definition: \\
Clarity: The style of an argumentation should be seen as clear if it uses grammatically correct and widely unambiguous language as well as if it avoids unnecessary complexity and deviation from the discussed issue. The used language should make it easy for you to understand without doubts what the author argues for and how. \\[2ex]

\noindent\#\#\# Question: \\
How would you rate the clarity of the style of the author's argumentation? Choose one of the options below \textcolor{gray}{[and explain your reasoning]}: \\
3 - High \\
2 - Medium \\
1 - Low \\
? - Cannot judge \\[1ex]
}\quad}

\caption{An expert prompt that contains instructions and an example issue, stance, and argument from the Dagstuhl-15512 ArgQuality corpus. This particular prompt example asks the model to rate the \textit{clarity} of the argument. The reasoning variant of this prompt is colored in gray.}
\label{expert-prompt}
\end{figure}

\subsection{Models}

Due to the complexity of the task, we focus on state-of-the-art LLMs for the automatic evaluation of argumentation quality. Building upon previous research regarding LLMs as annotators (cf. Section \ref{llm-as-annotators}), one of the most commonly used models is GPT-3 \cite{brown:2020}. Specifically, we use the \texttt{gpt-3.5-turbo-0613} accessible via OpenAI's API.\footnote{\tt \url{https://platform.openai.com/}} Despite the availability of the newer GPT-4 model \cite{openai:2023}, we do not employ it in our study due to the significantly higher associated costs.

In addition, we use Google's recently released PaLM 2 model \cite{anil:2023}, the successor to the original PaLM model \cite{chowdhery:2022}. The authors report comparable results to GPT-4 in semantic reasoning tasks, which makes it interesting for the evaluation of argument quality. For PaLM 2, we use the \texttt{text-bison@001} version of the model.

Both PaLM 2 and GPT-3 are closed-source language models. We initially intended to incorporate Meta's Llama 2 model \cite{touvron:2023} in our experiments, in order to evaluate the performance of open-source LLMs on our task. However, in pilot experiments, Llama 2 with 7 billion parameters did not follow the instructions and therefore did not produce quality scores. Even though the 13 billion parameter version of Llama 2 did generate quality scores, they were seemingly random, with agreement across the multiple runs close to zero. Due to hardware limitations, we did not test the largest Llama 2 model with 70 billion parameters. 

PaLM 2 and GPT-3 allow to specify a set of parameters such as temperature to control the diversity of the output, where lowering the temperature reduces the `randomness' of the output. For our experiments, we choose a reasonably low temperature of~0.3. Other parameters that we keep constant across models include $p=1.0$ of the nucleus sampling \cite{holtzman:2019}, most probable tokens $k=40$, and a maximum of 256 newly generated tokens.

\subsection{Prompting}
\label{prompting}

Two different groups of human annotators, the expert annotators of Wachsmuth et al. \cite{wachsmuth:2017} and the novice annotators recruited for this work, had access to different knowledge sources in their annotation guidelines. To determine whether the impact of this difference is similar between humans and LLMs, we created prompts that reflect the knowledge from the annotation guidelines of experts and novices. We refer to these prompt types as \textit{expert} and \textit{novice} prompts. 

Besides instructions, an expert prompt consists of an issue, a stance, and an argument from the Dagstuhl-15512-ArgQuality corpus. The expert prompt also contains the name and original definition of the quality dimension from Wachsmuth et al. \cite{wachsmuth:2017} as well as the annotation scheme (3-point Likert scale or ``not assessable''). An example of an expert prompt is shown in Figure \ref{expert-prompt}.

In contrast to the expert prompt type, novice prompts contain a conclusion (as described in Section~\ref{novice-annotation}) instead of an issue and stance. In the novice prompt, the definition of the quality dimension to be assessed is replaced by the simplified definition. However, the textual argument, which is renamed to ``reasons'', and the annotation scheme remain identical to the expert prompt. 

Recently, it has been shown that explanation-augmented prompts can elicit reasoning capabilities in LLMs and improve their performance across various tasks \cite{wei:2022,kojima:2022}.
In pilot experiments, we found that GPT-3 produces more consistent annotations if we prompt the model to provide an explanation for the chosen score. We therefore test \textit{reasoning} prompt variants in which we ask the model to provide an explanation for the generated quality rating. 

To take the randomness of the output of LLMs for the same prompt into account, each prompt variant is repeated (at least) three times for each argument and quality dimension. Each prompt repetition is considered as a separate annotator in order to calculate the agreement between the annotation and to draw conclusions about the consistency of the quality annotations of LLMs.

\section{Results}
\label{results}

To understand the strengths and weaknesses of large language models (LLMs) as argument quality assessors and to answer our research questions, we use the prompting approaches described in Section~\ref{prompting} to generate LLM annotations for arguments from the Dagstuhl-15512-ArgQuality corpus. The dataset contains 320~statements, 16~of which were originally judged as non-argumentative by expert human annotators and therefore are excluded from the analysis. 

\begin{figure}[t]
\centering
\includegraphics[trim=0 0.8cm 0 0, width=\textwidth]{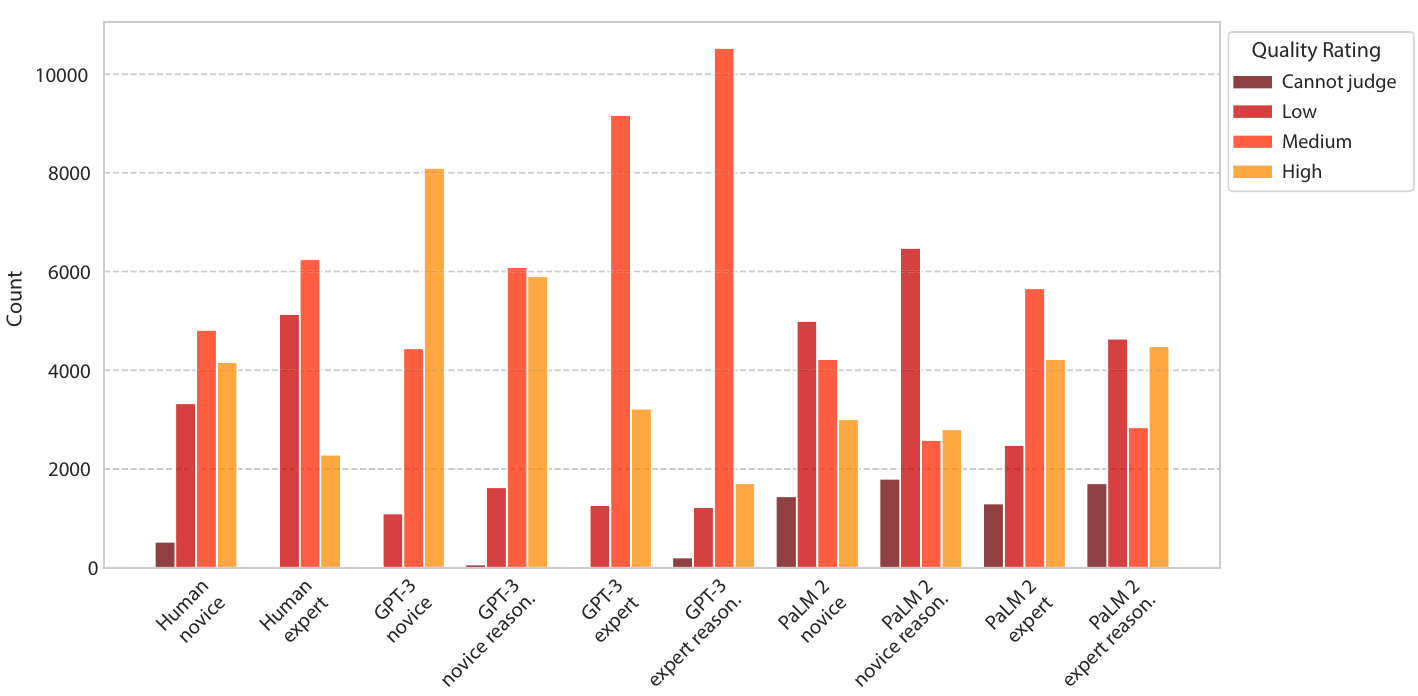}
\caption{Distribution of the assigned quality ratings across all quality dimensions compared between human annotators and LLMs.}
\label{rating-distribution-histogram}
\end{figure}

First, to identify biases in human and LLM argument quality annotations, we analyze the distribution of assigned labels across all quality dimensions. This distribution is visualized in Figure~\ref{rating-distribution-histogram}. Human annotations show an almost balanced distribution between low, medium and high quality ratings. However, it is noteworthy that human novices show a tendency to assign high ratings more frequently than experts. As for models, GPT-3 with expert prompts displays a much more skewed distribution, showing a strong bias towards medium ratings, deviating from the trend observed in human assessors. On the contrary, when GPT-3 is prompted with novice-level guidelines, it tends to assign high-quality ratings more frequently. Notably, annotations generated by PaLM~2 have a similar distribution to that of human annotators which seems promising for the subsequent analysis of agreement with human assessments.

Overall, it can be stated that not only the choice of model, but also the prompt type has a major influence on the generated argument quality ratings. Even slight prompt modifications, such as asking to justify the score, can result in a notable change in the assigned quality scores, which is especially prominent for PaLM~2 with expert prompts in our case. Another interesting observation is that GPT-3 almost always provides a rating for a given dimension: only in 214 out of 21,120 cases ($\approx$ 1\%) this model did not generate a score. The instances where PaLM~2 did not assess argument quality sum up to 4,972~($\approx$~23\%) and mostly stem from content policies, particularly in cases where arguments revolve around graphic topics such as pornography or contain offensive statements.

\subsection{Consistency of Argument Quality Annotations}

We address our first research question concerning the consistency of argument quality assessments by comparing the agreement levels within LLM groups with those of human assessors. To quantify the agreement within each group of annotators, we use Krippendorff's $\alpha$. To ensure a fair comparison with human annotators, we evaluate the agreement between three LLM annotation runs.

Table \ref{table-within-group-agreement} shows Krippendorff's $\alpha$ for human experts, human novices, and all LLM prompt variants across individual quality dimensions as well as overall agreement. Human annotators exhibit generally low agreement, with a maximum of 0.43 on the \textit{local acceptability} dimension for novices and 0.45 on \textit{reasonableness} for experts. This low level of agreement between humans emphasizes the subjectivity and complexity of assessing argument quality in a fine-grained taxonomy. For most of the quality dimensions, novice annotators show slightly higher agreement than those of experts, which could be due to the clearer definitions of the quality dimensions or perhaps due to the optimization of agreement for arguments that received more than three annotations (cf. Section \ref{novice-annotation}). 

In contrast, LLM agreement between annotation repetitions is substantially higher. Interestingly, the PaLM~2 model shows near-perfect agreement for both expert and novice prompts, but shows a notable drop when asked to explain its reasoning. In contrast to PaLM~2, the GPT-3 model exhibits a slight improvement in agreement when asked to provide an explanation. Such disparities might be due to the differences in the underlying architectures and training methodologies of the two models, which requires further exploration beyond the work at hand. Overall, both models show a high degree of agreement across different runs, with varying impact of reasoning prompts on the agreement depending on the employed model.

\paragraph{RQ1: Do LLMs provide more consistent evaluations of argument quality compared to human annotators?} The observed low agreement among human annotators underscores that evaluating argument quality is indeed a subjective and challenging task. In contrast, the significantly higher agreement among different LLM runs highlights the potential of these models for providing more consistent argument quality annotations.

\begin{table}[t]

\caption{Inter-annotator agreement per argument quality dimension within each group of human annotators and LLMs, reported as Krippendorf's $\alpha$. The dimension with the highest agreement within each group is marked in bold.}
\label{table-within-group-agreement}

\centering
\scriptsize
\setlength{\tabcolsep}{1.5pt}
\renewcommand{\arraystretch}{1.0}

\begin{tabular}{@{}l cc cc cc cc cc cc cc cc cc@{}}
\toprule
\bfseries Quality Dimension & \multicolumn{2}{c}{\bfseries Human} & \multicolumn{4}{c}{\bfseries GPT-3} & \multicolumn{4}{c}{\bfseries PaLM 2} \\
\cmidrule(l{\tabcolsep}r{\tabcolsep}){2-3}
\cmidrule(l{\tabcolsep}r{\tabcolsep}){4-7}
\cmidrule(l{\tabcolsep}r{\tabcolsep}){8-11}
& & & & & \multicolumn{2}{c}{\bfseries Reasoning} &  & & \multicolumn{2}{c}{\bfseries Reasoning} \\
\cmidrule(l{\tabcolsep}r{\tabcolsep}){6-7}
\cmidrule(l{\tabcolsep}r{\tabcolsep}){10-11}

& \multicolumn{1}{c}{Novice} & \multicolumn{1}{c}{Expert} & \multicolumn{1}{c}{Novice} & \multicolumn{1}{c}{Expert} & \multicolumn{1}{c}{Novice} & \multicolumn{1}{c}{Expert} & \multicolumn{1}{c}{Novice} & \multicolumn{1}{c}{Expert} & \multicolumn{1}{c}{Novice} & \multicolumn{1}{c}{Expert} \\

\arrayrulecolor{black}
\midrule
Cogency                 &	0.38 &	0.38 &	0.72 &	0.73 &	0.77 &	0.72 &\bf0.99&	0.98 &	0.73 &	0.74 \\
\ \ Local Acceptability &\bf0.43 &	0.33 &	0.64 &	0.69 &	0.70 &	0.75 &	0.98 &	0.97 &	0.60 &	0.71 \\
\ \ Local Relevance     &	0.36 &	0.41 &	0.70 &	0.59 &	0.76 &	0.61 &	0.98 &	0.98 &	0.78 &	0.68 \\
\ \ Local Sufficiency   &	0.35 &	0.27 &	0.74 &	0.69 &	0.79 &	0.72 &	0.98 &	0.97 &	0.63 &	0.63 \\
\addlinespace

Effectiveness           &	0.41 &	0.33 &	0.72 &	0.70 &	0.77 &	0.74 &	0.98 &\bf0.99 &	0.78 &	0.79 \\
\ \ Credibility         &	0.36 &	0.23 &\bf0.79&\bf0.79&	0.81 &	0.78 &\bf0.99&	0.97  &	0.72 &	0.67 \\
\ \ Emotional Appeal    &	0.35 &	0.21 &	0.73 &	0.56 &	0.74 &	0.70 &	0.98 &	0.97  &	0.64 &	0.72 \\
\ \ Clarity             &	0.27 &	0.25 &	0.72 &	0.69 &	0.71 &	0.69 &\bf0.99&\bf0.99 &\bf0.82&	0.80 \\
\ \ Appropriateness     &	0.39 &	0.17 &	0.66 &	0.50 &	0.68 &	0.58 &\bf0.99&\bf0.99 &	0.75 &	0.81 \\
\ \ Arrangement         &	0.39 &	0.26 &	0.68 &	0.66 &	0.71 &	0.69 &\bf0.99&\bf0.99 &	0.69 &	0.65 \\
\addlinespace

Reasonableness          &	0.35 &\bf0.45 &	0.73 &	0.78 &	0.81 &	0.74 &	0.97 &	0.97 &	0.70 &	0.76 \\
\ \ Global Acceptability&   0.37 &	0.39 &	0.72 &	0.77 &	0.77 &	0.74 &	0.98 &	0.97 &	0.66 &	0.70 \\
\ \ Global Relevance    &	0.38 &	0.26 &	0.69 &	0.71 &	0.81 &	0.70 &\bf0.99&	0.98 &	0.74 &\bf0.85 \\
\ \ Global Sufficiency  &	0.27 &	0.17 &	0.72 &	0.69 &	0.72 &	0.75 &	0.98 &	0.96 &	0.62 &	0.47 \\
\addlinespace

Overall Quality         &	0.41 &	0.44 &	0.77 &	0.77 &\bf0.82 &\bf0.81&	0.98 &	0.97 &	0.77 &	0.78 \\
\midrule
\bfseries Overall $\alpha$    &	0.37 &	0.40 &	0.76 &	0.73 &	0.78 &	0.74 &	0.99 &	0.98 &	0.76 &	0.78 \\
\bottomrule

\end{tabular}
\end{table}

\subsection{Agreement between Humans and LLMs}

\begin{table}[t]

\centering
\scriptsize
\setlength{\tabcolsep}{2pt}
\renewcommand{\arraystretch}{1.1}

\caption{
Number of arguments with perfect agreement for each argument dimension within each group of human annotators (expert, novice).
}
\label{table-perfect-agreement}

\centering
\begin{tabular}{@{}l c c @{}}
\toprule
\bfseries Quality Dimension & \bf Expert & \bf Novice \\           

\arrayrulecolor{black}
\midrule
Cogency                 & 122               & 105 \\
\ \ Local Acceptability & \phantom{0}82     & 115 \\
\ \ Local Relevance     & \phantom{0}99     & 100 \\
\ \ Local Sufficiency   & 113               & \phantom{0}89 \\
\addlinespace
Effectiveness           & 128               & 118 \\
\ \ Credibility         & 115               & \phantom{0}86 \\
\ \ Emotional Appeal    & 130               & \phantom{0}90 \\
\ \ Clarity             & \phantom{0}89     & \phantom{0}92 \\
\ \ Appropriateness     & \phantom{0}53     & 102 \\
\ \ Arrangement         & \phantom{0}81     & 102 \\
\addlinespace
Reasonableness          & 126               & 119 \\
\ \ Global Acceptability& \phantom{0}96     & 102 \\
\ \ Global Relevance    & \phantom{0}66     & \phantom{0}96 \\
\ \ Global Sufficiency  & 136               & \phantom{0}81 \\
\addlinespace
Overall Quality         & 134               & 130 \\

\bottomrule
\end{tabular}
\end{table}

\begin{figure}[t]
\centering
\includegraphics[width=\textwidth]{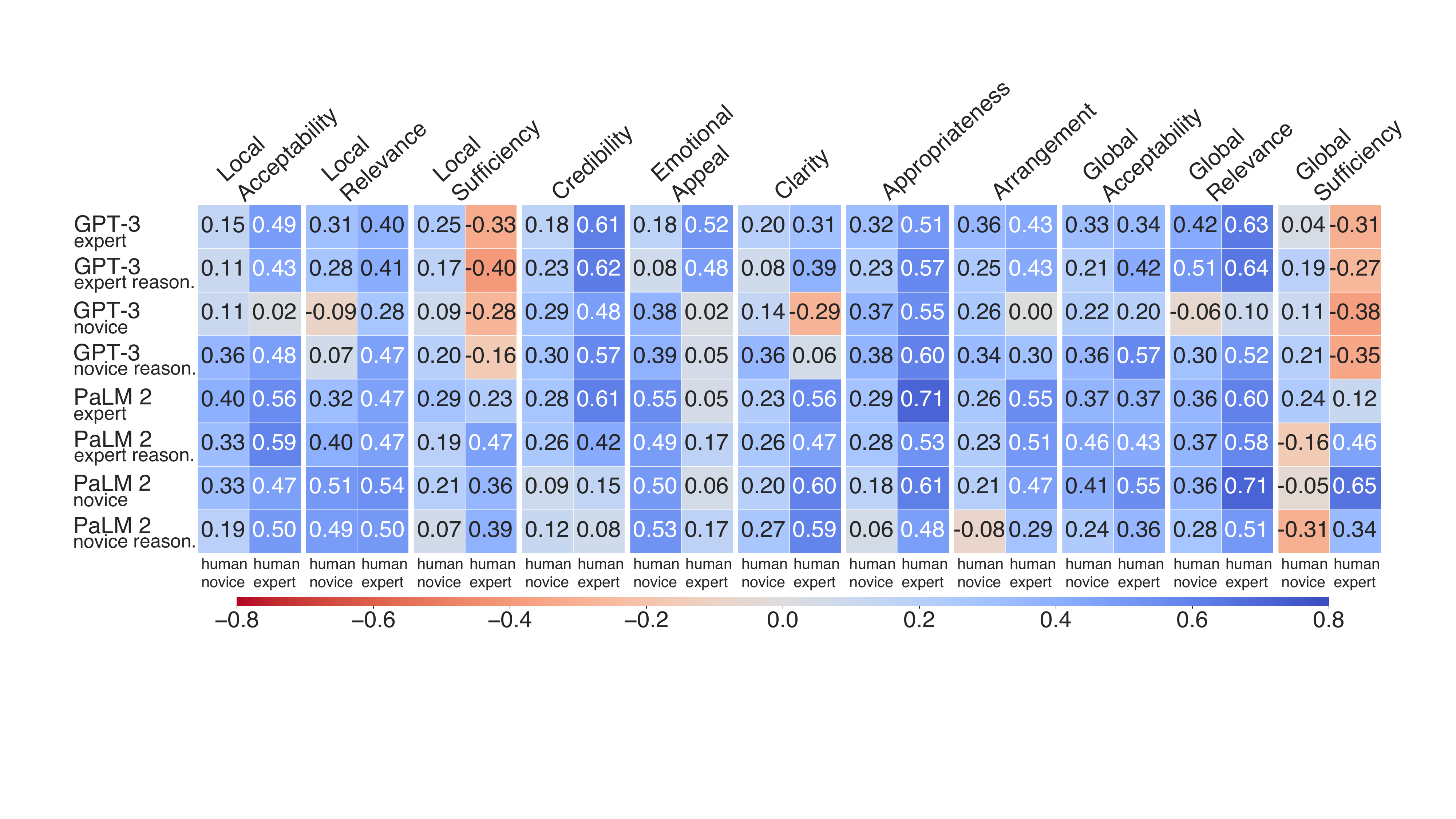}
\caption{Inter-annotator agreement (Krippendorff's $\alpha$) between human and LLM annotations for each fine-grained argument quality dimension.}
\label{fine-grained-agreement-heatmap}
\end{figure}

\begin{figure}[!t]
\begin{minipage}{0.48\textwidth}
\centering
\includegraphics[width=\textwidth]{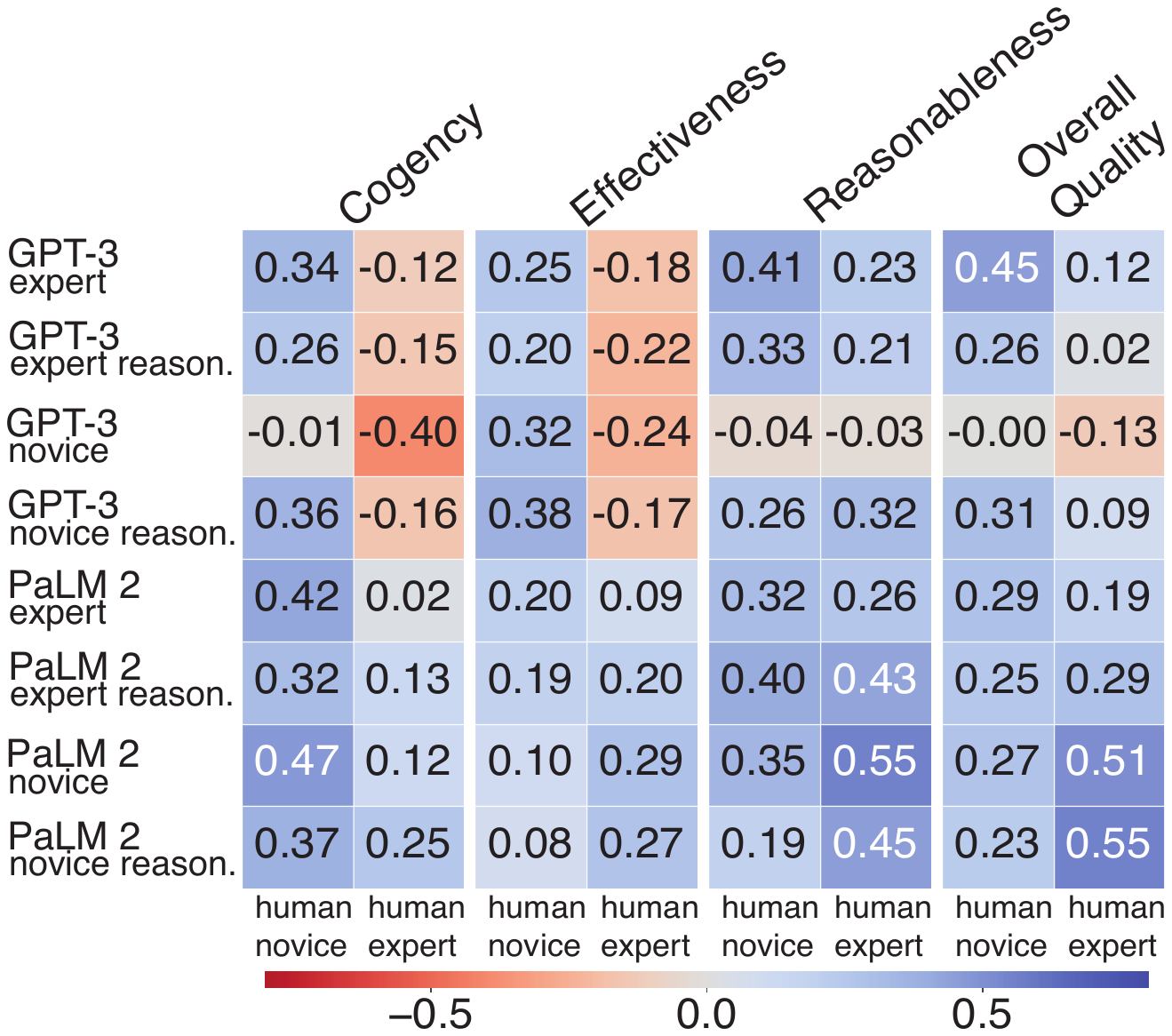}
\caption{Inter-annotator agreement (Krippendorff's $\alpha$) between human and LLM annotations for each coarse-grained argument quality dimension.}
\label{coarse-grained-agreement-heatmap}
\end{minipage}\hfill
\begin{minipage}{0.48\textwidth}
\centering
\vspace*{0.8cm}
\includegraphics[width=\textwidth]{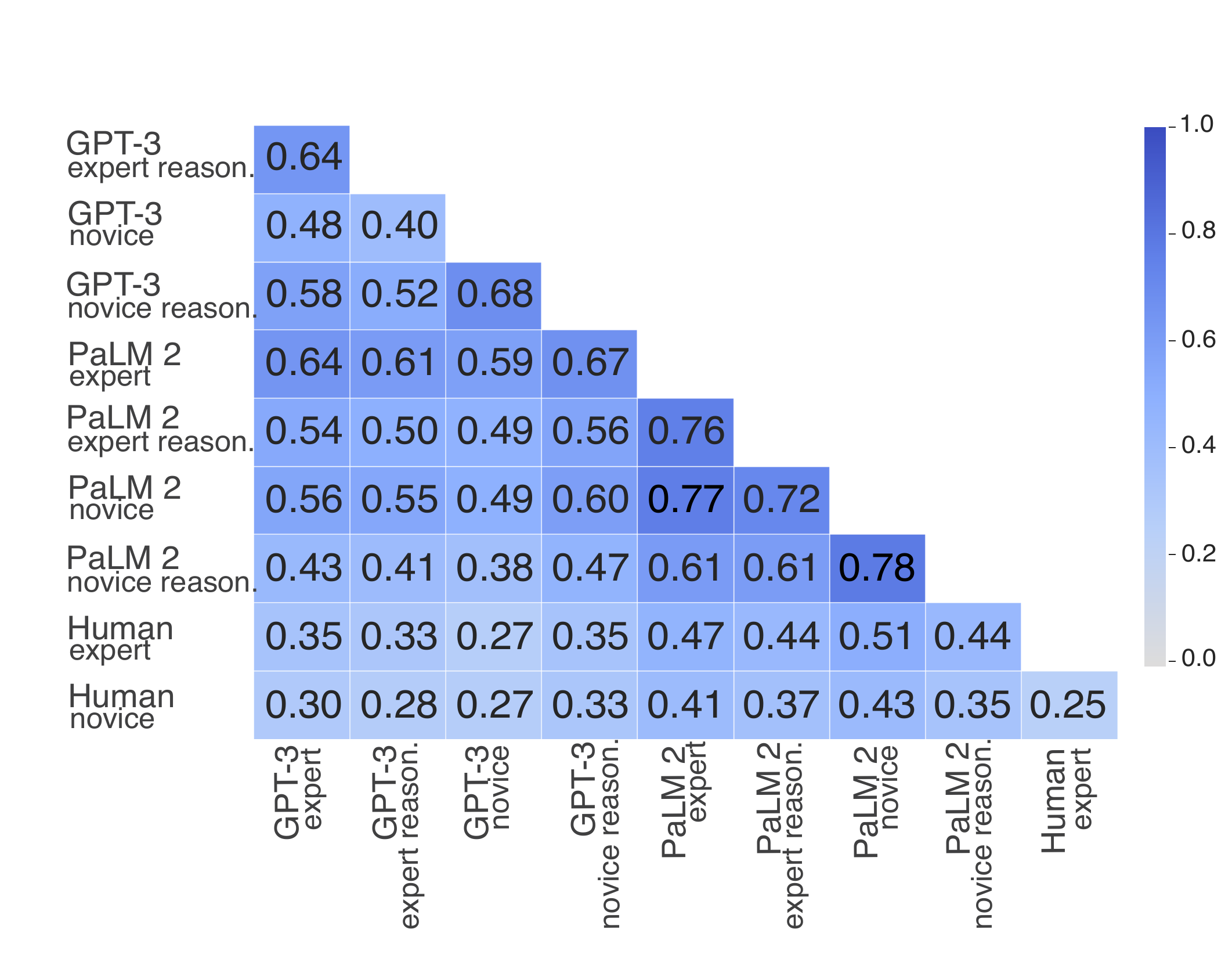}
\caption{Overall inter-annotator agreement (Krippendorff's $\alpha$) between each combination of human expert, novice, and LLM-generated annotations.}
\label{intra-group-agreement-heatmap}
\end{minipage}
\end{figure}

We discovered that LLMs generate annotations more consistently than humans. However, to assert that LLMs can reliably evaluate the quality of arguments, we need to test how the automatic annotations align with the human annotations. Given the low agreement among human annotators, we created subsets of arguments for each quality dimension, where either all expert annotators or all novice annotators unanimously agreed on a score. Table~\ref{table-perfect-agreement} presents the statistics of the resulting subsets with perfect agreement, which we employ for further inter-annotator agreement analysis. 

Figure~\ref{fine-grained-agreement-heatmap} shows the agreement for each quality dimension, as measured by Krippendorff's $\alpha$, between human annotations and automatically generated quality ratings by LLMs with different prompts. Overall, we observe moderate agreement across most quality dimensions, with the annotations by PaLM~2 reaching a maximum of 0.71 for \textit{appropriateness} and \textit{global relevance}. Regardless of the prompt type, PaLM~2 annotations generally achieve higher agreement with human annotations compared to GPT-3. In the case of \textit{local} and \textit{global sufficiency}, there are even systematic disagreements between the GPT-3 assessments and those of human experts. Similarly, disagreement is observed between PaLM~2 annotations and human novices for the \textit{global sufficiency} dimension. 

Overall, there is a large variance in agreement between model and human judgments across different quality dimensions. For example, while the agreement on \textit{credibility} and \textit{appropriateness} are in the range of [0.08, 0.62] and [0.06, 0.71] respectively, the agreement on \textit{local} and \textit{global sufficiency} fluctuates even more.

In terms of prompt variants, we can see that GPT-3 with expert prompts shows a higher agreement with human expert annotations than with human novice annotations, and a similar trend is observed for GPT-3 with novice prompts and human novices. On the other hand, PaLM~2 with either of the prompt types tends to show higher agreement with human experts. Similar findings can be inferred from the agreement between LLMs and human novices and experts on the coarse-grained quality dimensions that are visualized in Figure~\ref{coarse-grained-agreement-heatmap}.

\paragraph{RQ2: Do the assessments of argument quality made by LLMs align with those made by either human experts or human novices?} We found that LLMs agree most with human argument quality ratings on fine-grained quality dimensions such as \textit{credibility}, \textit{emotional appeal}, \textit{appropriateness}, and \textit{global relevance} or on coarse-grained dimensions such as \textit{reasonableness} and \textit{overall quality}. Overall, we found varying degrees of agreement between LLMs and human annotators, with PaLM~2 annotations tending to generally align more with those of humans.

\subsection{LLMs as Additional Annotators}

LLMs can be employed either as independent automatic argument quality raters or as a source of additional annotations to validate a set of human annotations. For the second scenario, we analyze the overall agreement between different combinations of human (expert or novice) and LLM annotators. 

Figure~\ref{intra-group-agreement-heatmap} illustrates the overall Krippendorff's $\alpha$ agreement for each combination of annotator group. We can see that there is low to medium agreement for each combination of annotators, with the lowest value being 0.25 between human novices and human experts and the highest value being 0.77 between PaLM with novice and expert prompts. Regardless of the prompt type, the agreement between PaLM 2 and GPT-3 is moderate, ranging from 0.38 to 0.67. This suggests the potential efficacy of employing diverse models as supplementary annotators.

We further investigate whether the agreement changes if we incrementally integrate automatically generated annotations to the original set of human annotations. The results reported in Table~\ref{table-agreement-model-runs} show that adding PaLM~2 annotations can significantly improve the agreement of human experts as well as human novices. A significant increase is already visible after adding three annotations to the annotations of human experts and four to the annotations of human novices. However, the introduction of GPT-3 annotations leads to a significant decrease in agreement. This can be attributed to the relatively low level of agreement between GPT-3 and human annotators (cf. Figure \ref{intra-group-agreement-heatmap}).

\begin{table}[t]
\centering
\scriptsize
\setlength{\tabcolsep}{1.2pt}
\renewcommand{\arraystretch}{1.1}
\caption{Change in overall Krippendorf's $\alpha$ after adding LLM annotations to human expert or novice annotations. Significant changes ($p < 0.05$) between the agreement of the original annotations and the modified annotations set are marked with *.}
\label{table-agreement-model-runs}
\begin{minipage}{.5\linewidth}
\centering
\begin{tabular}{@{}l c c c c @{}}
\toprule

\bf Annotations & \multicolumn{2}{c}{\bf Expert} & \multicolumn{2}{c}{\bf Novice} \\
\cmidrule(l{\tabcolsep}r{\tabcolsep}){2-3}
\cmidrule(l{\tabcolsep}r{\tabcolsep}){4-5}
& \multicolumn{1}{c}{GPT-3} & \multicolumn{1}{c}{PaLM 2} & \multicolumn{1}{c}{GPT-3} & \multicolumn{1}{c}{PaLM 2} \\

\arrayrulecolor{black}
\midrule
Human experts         & 0.40\phantom{$^*$}      & 0.40\phantom{$^*$}& 0.40\phantom{$^*$}& 0.40\phantom{$^*$}  \\
\addlinespace
+1 annotation         & 0.32$^*$                & 0.37$^*$          & 0.30$^*$          & 0.37$^*$  \\
+2 annotations        & 0.31$^*$                & 0.40\phantom{$^*$}& 0.32$^*$          & 0.40\phantom{$^*$}  \\
+3 annotations        & 0.33\phantom{$^*$}      & 0.44$^*$          & 0.35\phantom{$^*$}& 0.44$^*$  \\
+4 annotations        & 0.35\phantom{$^*$}      & 0.47$^*$          & 0.39\phantom{$^*$}& 0.47$^*$  \\
+5 annotations        & 0.37\phantom{$^*$}      & 0.50$^*$          & 0.42$^*$          & 0.50$^*$  \\
\bottomrule
\end{tabular}
\end{minipage}%
\begin{minipage}{.5\linewidth}
\centering
\begin{tabular}{@{}l c c c c @{}}
\toprule
\bf Annotations & \multicolumn{2}{c}{\bf Expert} & \multicolumn{2}{c}{\bf Novice} \\
\cmidrule(l{\tabcolsep}r{\tabcolsep}){2-3}
\cmidrule(l{\tabcolsep}r{\tabcolsep}){4-5}            
& \multicolumn{1}{c}{GPT-3} & \multicolumn{1}{c}{PaLM 2} & \multicolumn{1}{c}{GPT-3} & \multicolumn{1}{c}{PaLM 2} \\

\arrayrulecolor{black}
\midrule
Human novices         & 0.37\phantom{$^*$} & 0.37\phantom{$^*$}   & 0.37\phantom{$^*$}& 0.37\phantom{$^*$}  \\
\addlinespace
+1 annotation         & 0.27$^*$           & 0.29$^*$             & 0.27$^*$          & 0.27$^*$  \\
+2 annotations        & 0.26$^*$           & 0.33\phantom{$^*$}   & 0.29$^*$          & 0.30\phantom{$^*$}   \\
+3 annotations        & 0.28$^*$           & 0.37$^*$             & 0.33\phantom{$^*$}& 0.35$^*$  \\
+4 annotations        & 0.30$^*$           & 0.41$^*$             & 0.37\phantom{$^*$}& 0.39$^*$  \\
+5 annotations        & 0.32$^*$           & 0.45$^*$             & 0.40$^*$          & 0.43$^*$  \\
\bottomrule
\end{tabular}
\end{minipage} 
\end{table}

\paragraph{RQ3: Can integrating LLM annotations with human annotations significantly improve the resulting agreement in argument quality ratings?} 

The analysis indicates that the impact on agreement levels when incorporating generated quality assessments with human annotations varies based on the employed LLM. When using a powerful model such as PaLM~2, the agreement of human annotations can be significantly increased by adding three or more generated annotations. These results underscore LLMs as valuable contributors to the annotator ensemble.

\section{Conclusion}
\label{conclusion}

In this paper, we investigated the effectiveness of LLMs, specifically GPT-3 and PaLM 2, in evaluating argument quality. We utilized four distinct prompt types to solicit quality ratings from these models and compared their assessments with those made by human novices and experts. The results reveal that LLMs exhibit greater consistency in evaluating argument quality compared to both novice and expert human annotators, showcasing their potential reliability. Based on our empirical analysis, we can recommend two modes of application for LLMs as annotators of argument quality: (1) a fully automatic annotation procedure with LLMs as automatic quality raters, for which we found moderately high agreement between PaLM~2 and human expert quality ratings, or (2) a semi-automatic procedure using LLMs as additional quality annotators, resulting in a significant enhancement in agreement when combined with human annotations. In both modi, LLMs can serve as a valuable tool for streamlining the argument quality annotation process on a large scale.

To further minimize annotation expenses, we intend to expand these experiments to various open-source large language models. In addition to the investigated zero-shot prompting technique, enhancing agreement with human annotations could involve utilizing few-shot prompting technique or fine-tuning LLMs based on human judgments of argument quality. We see great potential in LLMs as argument quality raters, which, if further optimized to agree more closely with human assessments, can reduce manual effort and expenses, establishing them as valuable tools in argument mining.

\section{Limitations}

The experiments in this paper are based on the hypothesis that multiple runs of the same model, prompt, and hyperparameters simulate different annotators as a result of nucleus sampling. This hypothesis has not yet been proven, and its validity cannot be inferred from the analysis. The higher inter-model agreement indicates a lower variance in the automatically generated annotations, which might argue against this hypothesis. Therefore, the agreement between model and human annotations has been calculated using examples with perfect agreement only in order to exclude effects of this variance. 
Further experiments are needed to determine how to replicate the behavior of different annotators. 

Although we deeply investigate LLMs as quality assessors for arguments, the generalizability of our results beyond argumentation is not yet clear. However, due to the complexity and subjectivity of argument quality assessment, as seen from low human inter-annotator agreement, we argue that this task might be a worst-case scenario for LLMs, and we would expect comparable or even better results in less subjective task domains.  However, more experiments are needed to confirm or reject this hypothesis.

\subsubsection{Acknowledgements}
This research has been supported by the German Research Foundation (DFG) with grant number 455911521, project ``LARGA'' in SPP ``RATIO''.

\bibliographystyle{splncs04}
\bibliography{ratio24-lit}

\newpage
\appendix{
\section*{Appendix}
\label{appendix}
\begin{table}[!ht]
\fontsize{8pt}{10pt}\selectfont
\setlength{\tabcolsep}{4pt}
\centering
\caption{The set of simplified definitions of argument quality dimensions. }
\label{table-simplified-definitions}
\begin{tabular}{@{}l p{9cm}@{}}
\toprule
\bf Quality Dimension & \bf Definition\\
\midrule
Local acceptability & The reasons are individually believable: they could be true.\\
Local relevance & The reasons (assuming they are true) are relevant to the conclusion: they tell why one could accept the conclusion. \\
Local sufficiency & The reasons (assuming they are true) are sufficient to draw the conclusion: no other reason is necessary to arrive at the conclusion.\\
\addlinespace
Credibility & The reasons make the author seem trustworthy and knowledgeable.\\
Emotional appeal & The reasons can make people feel emotions that make them willing to agree with the author.\\ 
Clarity & The author uses clear, grammatically correct and unambiguous language. The author sticks to the main topic and does not make things overly complicated. \\ 
Appropriatness & The author uses an appropriate style for the reasons and this style fits to the topic's importance.\\
Arrangement & The reasons are properly organized: coherent, easy to follow, well-structured. \\
\addlinespace
Global acceptability & The reasons and conclusion combined are believable: everything taken together could be true.\\
Global relevance & The reasons (assuming they are true) are relevant for resolving a discussion around the conclusion's topic. \\
Global sufficiency & The reasons (assuming they are true) are sufficient in that they consider any counter-arguments that may arise.\\
\bottomrule
\end{tabular}
\end{table}
}

\end{document}